\documentclass[lettersize,journal]{IEEEtran}
\usepackage{amsmath,amsfonts}
\usepackage{algorithmic}
\usepackage{algorithm}
\usepackage{array}
\usepackage[caption=false,font=normalsize,labelfont=sf,textfont=sf]{subfig}
\usepackage{textcomp}
\usepackage{stfloats}
\usepackage{url}
\usepackage{verbatim}
\usepackage{graphicx}
\usepackage{cite}
\begin{document}

\title{Real-World Image Super Resolution via Unsupervised Bi-directional Cycle Domain Transfer Learning based Generative Adversarial Network}

\author{Xiang Wang,
        Yimin Yang,~\IEEEmembership{Senior Member, IEEE,} Zhichang Guo, Zhili Zhou,~\IEEEmembership{Member,~IEEE}, Yu Liu,~\IEEEmembership{Member,~IEEE}, Qixiang Pang, Shan Du,~\IEEEmembership{Senior Member, IEEE}
\thanks{X. Wang is with the Dept. of Computer Science, Lakehead University, Thunder Bay, ON, Canada, e-mail: xwang150@lakeheadu.ca}
\thanks{Y. Yang is with the Dept. of electrical and computer engineering, Western University, London, ON, Canada, e-mail: yyan2294@uwo.ca}

\thanks{Z. Guo is with the School of Mathematics, Harbin Institute of Technology, Hei Long Jiang, China, e-mail: mathgzc@hit.edu.cn}

\thanks{Z. Zhou is with the School of Computer and Software, Nanjing University of Information Science and Technology, Jiangsu, China, zhou\_zhili@163.com}

\thanks{Y. Liu is with the School of Microelectronics, Tianjin University, Tianjin, China, e-mail: liuyu@tju.edu.cn}

\thanks{Q. Pang is with the Okanagan College, Kelowna, BC, Canada, e-mail: kpang@okanagan.bc.ca}

\thanks{S. Du is with the Dept. of Comp. Sci., Math, Physics \& Statistics, The University of British Columbia Okanagan, BC, Canada, e-mail: shan.du@ubc.ca}
}
\markboth{Journal of \LaTeX\ Class Files,~Vol.~14, No.~8, August~2021}%
{Shell \MakeLowercase{\textit{et al.}}: A Sample Article Using IEEEtran.cls for IEEE Journals}


\maketitle

\begin{abstract}
Deep Convolutional Neural Networks (DCNNs) have exhibited impressive performance on image super-resolution tasks. However, these deep learning-based super-resolution methods perform poorly in real-world super-resolution tasks, where the paired high-resolution and low-resolution images are unavailable and the low-resolution images are degraded by complicated and unknown kernels. To break these limitations, we propose the Unsupervised Bi-directional Cycle Domain Transfer Learning-based Generative Adversarial Network (UBCDTL-GAN), which consists of an Unsupervised Bi-directional Cycle Domain Transfer Network (UBCDTN) and the Semantic Encoder guided Super Resolution Network (SESRN). First, the UBCDTN is able to produce an approximated real-like LR image through transferring the LR image from an artificially degraded domain to the real-world LR image domain. Second, the SESRN has the ability to super-resolve the approximated real-like LR image to a photo-realistic HR image. Extensive experiments on unpaired real-world image benchmark datasets demonstrate that the proposed method achieves superior performance compared to state-of-the-art methods.
\end{abstract}


\begin{figure*}[t!]
\centerline{\includegraphics[width =\textwidth
]{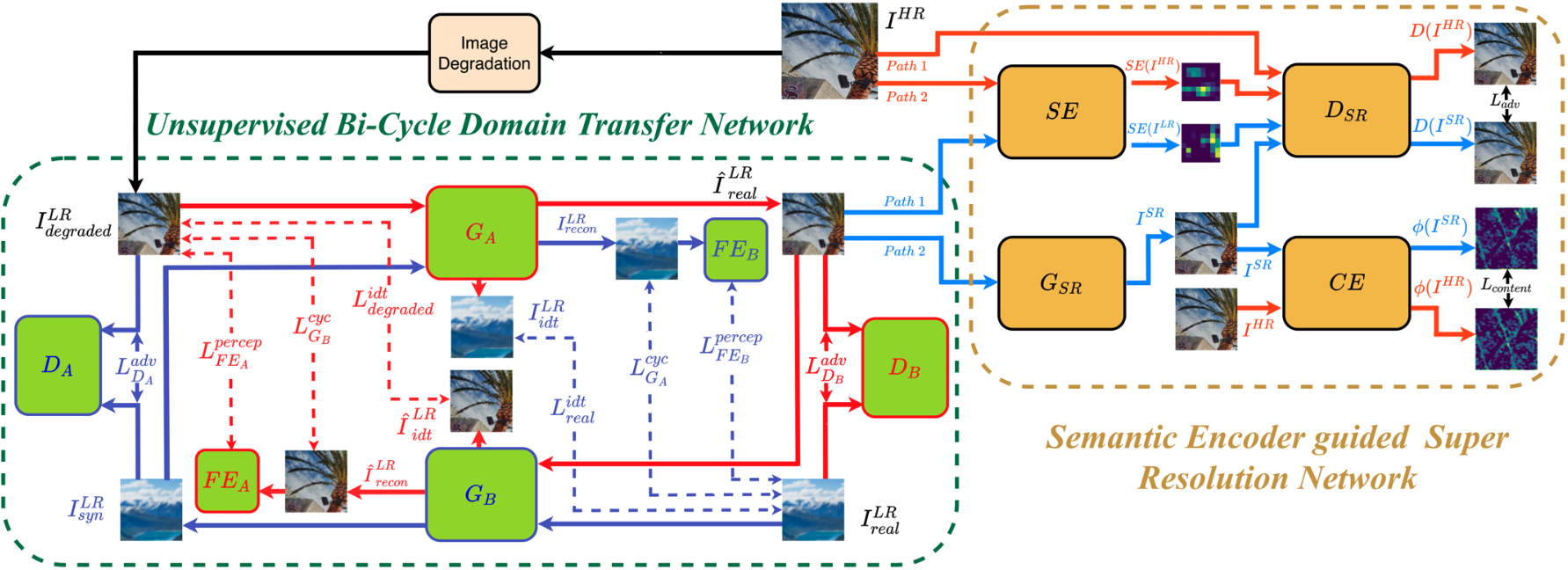}}
\caption{\textbf{The overview of the proposed UBCDTL-GAN}: In the first stage (left), the green dot rectangle represents the Unsupervised Bi-directional Cycle Domain Transfer Network (UBCDTN). The red path indicates the forward cycle module. Given the input HR image $I^{HR}$, $I^{LR}_{degraded}$ is the artificially degraded LR image, and $\widehat{I}^{LR}_{real}$ is the real-like LR image generated by a generator $G_{A}$. $\widehat{I}^{LR}_{recon}$ represents the reconstructed image of $I^{LR}_{degraded}$ produced by another generator $G_{B}$ from $\widehat{I}^{LR}_{real}$. $\widehat{I}^{LR}_{idt}$ is produced by $G_{B}$ from the original $I^{LR}_{degraded}$. In addition, $L^{adv}_{D_{B}}$, $L^{idt}_{degraded}$, $L^{cyc}_{G_{B}}$ and $L^{percep}_{FE_{A}}$ depicted in red dotted line represent adversarial loss, identity loss, cycle-consistency loss and cycle-perceptual loss for forward cycle module. $D_{A}$ is a discriminator and $FE_{A}$ denotes a feature extractor. Symmetrically, the blue path shows the backward cycle module, where $I^{LR}_{real}$ is given by a real-world dataset and synthesized LR image $I^{LR}_{syn}$ is generated by $G_{B}$. Moreover, the $G_{A}$ is able to translate $I^{LR}_{syn}$ back to reconstructed real-world LR image $L^{LR}_{recon}$ and generate the identity real-world LR image  $I^{LR}_{idt}$. The blue dotted line represents the adversarial loss $L^{adv}_{D_{A}}$, identity loss $L^{idt}_{real}$, cycle-consistency loss $L^{cyc}_{G_{A}}$ and cycle-perceptual loss $L^{percept}_{FE_{B}}$ for backward cycle module respectively. $D_{B}$ is a discriminator and $FE_{B}$ denotes a feature extractor. In the second stage (right), the framework of a Semantic Encoder guided Super Resolution Network (SESRN) is depicted in a yellow dot rectangle, where it consists of Semantic Encoder $SE$, Generator $G_{SR}$, Joint Discriminator $D_{SR}$ and Content Extractor $\phi$. There are two paths in the SESRN, where the red path indicates the process of the real tuple and the blue path is the process of the fake tuple. $I^{SR}$ is the SR image from $G_{SR}$. Furthermore, $SE(\cdot)$ denotes the embedded semantics obtained from $SE$. $D(\cdot)$ represents the output probability of $D_{SR}$. $\phi(\cdot)$ describes the features learned by  $\phi$.}
\end{figure*}

\section{Introduction}
\IEEEPARstart{S}{ingle} Image Super-Resolution (SISR) aims to reconstruct a High-Resolution (HR) image from a single Low-Resolution (LR) image. It has been widely applied in many computer vision applications, such as surveillance \cite{UL_surveillance}, image enhancement \cite{UL_image_enhancement} and medical image processing \cite{UL_medical_image_processing}. In the SISR task, the general degradation formula is expressed as:
{
\begin{equation}
y=(x\bigotimes k)\downarrow s + n
\end{equation}}where $x$ represents the HR image, $y$ is the degraded LR image, $k$ denotes a blur kernel, and $\bigotimes$ is the convolution operation performed on $x$ and $k$. $\downarrow s$ denotes a downsampling operation on the image with a scaling factor $s$, and $n$ is considered as an additive white Gaussian noise. However, under the real-world scene setting, $k$ is unknown and $n$ should take into account many possible conditions such as sensor noise, compression artifacts, and unpredicted noise caused by physical devices. Resulting from the existence of uncertain kernel $k$ and noise $n$, SISR has become a particularly ill-posed inverse task since there are infinite HR images that can be recovered from a given LR image, in which it is required to select the most plausible solution.

Recently, a great number of deep learning-based SISR models have been proposed, such as RDN \cite{RDN}, EDSR \cite{EDSR}, SRDenseNet \cite{SRDenseNet}, and ESRGAN \cite{ESRGAN_Wang}. Despite the successful progress achieved by the aforementioned methods, there still exist unnoticed issues. Those models were trained in a supervised manner with a large number of paired images, synthesized LR images produced by pre-determined degradation and its HR counterpart, resulting in deteriorative performance when they are applied to real-world scenarios. The reason is that paired LR-HR data is unavailable and the degradation of the input LR image is unknown in the real-world scene. Besides, it is unreasonable to simply apply the LR images downsampled by the ideally fixed kernel to the real-world SR problems \cite{kim2020unsupervised,maeda2020unpaired}. 
There exists a large domain gap between real-world LR images and artificially synthesized LR images. In addition, the synthesized LR images may eliminate diverse patterns and complicated characteristics belonging to real-world LR images such as sensor noise and natural characteristics. Thus, the existing SR methods normally encounter a serious domain consistency problem and produce poor performance in practical scenarios \cite{CinCGAN,zhao2018unsupervised}.

Therefore, it is imperative to explore an effective method that can apply unpaired images to satisfy the need for real-world SR scenarios. It must be different from the aforementioned SR methods which do not take into consideration the domain gap between the LR images generated from a known degradation (e.g., bicubic downscaling) and the real-world LR images. To address the above limitations, we propose an Unsupervised Bi-directional Cycle Domain Transfer Learning-based Generative Adversarial Network (UBCDTL-GAN) for real-world image super-resolution, which is composed of two networks, Unsupervised Bi-directional Cycle Domain Transfer Network (UBCDTN) and Semantic Encoder guided Super Resolution Network (SESRN). To simulate the real-world data distribution and reduce the domain gap between the generated LR image domain and the real-world LR image domain, we design UBCDTN to estimate the inherent degradation kernel from the real-world LR distribution and translate the artificially degraded LR domain image to the real-world domain. With the help of the cycle consistency mechanism \cite{CycleGAN}, the proposed UBCDTN is able to learn bi-directional inverse mapping in an unsupervised manner, which can ensure the generated real-like LR image preserves desired characteristics of real-world patterns. Besides, we also enforce auxiliary constraints on the UBCDTN such as adversarial loss, identity loss, and cycle-perceptual loss. The designed domain transfer network provides an effective way to generate real-like LR images, which can construct the paired real-world LR-HR data for the following SESRN. For the second step, we design the Semantic Encoder guided Super Resolution Network (SESRN). The goal of the SESRN is to super-resolve the real-like LR images to the photo-realistic HR images. It is emphasized that we apply our previously proposed architecture and optimization strategy of GAFHH-RIDN \cite{xiang2021end-to-end} to the generator and discriminator of SESEN. We evaluate our method on the NTIRE 2020 Super Resolution Challenge Track 1 validation dataset. The quantitative and qualitative comparisons demonstrate the superiority of the proposed UBCDTL-GAN compared with other state-of-the-art methods. A comparison of visual results with various latest methods is shown in Fig. 4, and the numeric results can be seen in Table I. The main contributions of our proposed method can be summarized as follows:

1) We propose a novel bi-directional cycle domain transfer network, UBCDTN. According to the domain transfer learning scheme, the designed bi-directional cycle architecture is able to eliminate the domain gap between the artificially degraded LR images and real-world LR images in an unsupervised manner.

2) We further impose the auxiliary constraints on UBCDTN by incorporating adversarial loss, identity loss, and cycle-perceptual loss, which can guarantee that the generated real-like LR images contain the same style as real-world images.

3) We design the SESRN as a deep super-resolution network to generate visually pleasant SR results under supervised learning settings.

4) Benefiting from the collaborative training strategy, the proposed UBCDTL-GAN is able to be trained in an end-to-end fashion, which can ease the entire training procedure and strengthen the robustness of the model.

\begin{figure*}[t!]
\centerline{\includegraphics[width =\textwidth]{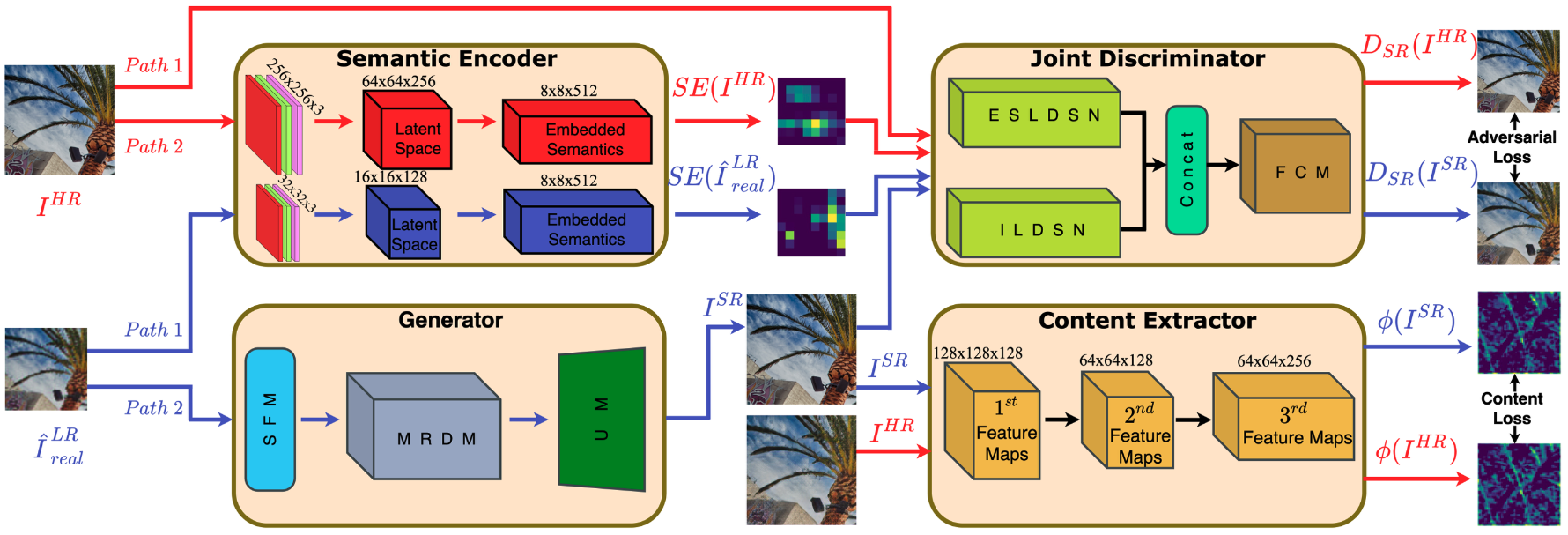}}
\caption{\textbf{The proposed SESRN and its components}: Semantic Encoder $SE$, Generator $G_{SR}$, Joint Discriminator $D_{SR}$ and Content Extractor $\phi$. For $D_{SR}$, ESLDSN represents the Embedded Semantics-Level Discriminative Sub-Net, ILDSN represents the Image-Level Discriminative Sub-Net, and FCM denotes the Fully Connected Module. As for the generator $G_{SR}$, there are three stages: Shallow Feature Module (SFM), Multi-level Dense Block Module (MDBM), and Upsampling Module (UM). $I^{HR}$ and $\widehat{I}^{LR}_{real}$ denote HR images and LR images respectively. $I^{SR}$ is SR images from $G_{SR}$. Furthermore, $SE(\cdot)$ denotes the embedded semantics obtained from the $SE$. $D_{SR}(\cdot)$ represents the output probability of the $D_{SR}$. $\phi(I^{HR})$ and $\phi(I^{SR})$ describes the features learned by the content extractor $\phi$.}
\label{fig}
\end{figure*}
\section{Related Work}
In this section, we first review the paired image super-resolution methods. Then, we introduce blind and unsupervised learning methods for real-world scenarios.
\subsection{Paired Image Super Resolution}
In recent years, deep learning-based methods have exhibited exceptional ability to enhance SISR performance. However, most of these methods rely on supervised settings, where specific pre-defined and paired LR and HR images are required in the training process. The pioneer method was introduced by Dong $et \ al.$ \cite{SRCNN}, namely SRCNN. It is able to recover HR images from bicubic downsampled LR images in an end-to-end manner. Moreover, the enhanced deep SR method EDSR \cite{EDSR} was proposed by Lim $et\ al.$, which greatly enlarges the network to be deeper and wider, resulting in learning richer feature representations. Later, generative adversarial learning was introduced in SISR. SRGAN \cite{SRGAN} proposed by Ledig $et\ al.$ incorporates adversarial loss and perceptual loss to update the generator and discriminator respectively, which can obtain photo-realistic SR images. Furthermore, the ESRGAN \cite{ESRGAN_Wang} introduces the new generator architecture and adopts enhanced adversarial loss, achieving promising performance in the visual aspect.

\subsection{Blind Image Super Resolution}
The blind SISR is defined as the task that supposes the LR image is produced by an unknown degradation kernel from its HR version. The Iterative Kernel Correction (IKC) \cite{IKC} was proposed by Gu $et\ al$. to estimate blur kernel and eliminate artifacts caused by kernel discrepancy. Zhang $et\ al$. \cite{IRCNN} proposed IRCNN which includes a set of CNN denoisers to estimate the blur model. However, the results of IRCNN indicate that it is difficult to estimate a comprehensive degradation kernel in real-world conditions. The blind SR methods have the limited capability to approximate unknown degradations. Thus, there is still room for improvement in dealing with blind image super-resolution.

\subsection{Unpaired Super Resolution}
Recently, many unsupervised methods have been proposed to satisfy real-world conditions where the paired LR-HR image data is unavailable, and the input image is corrupted by unknown degradation kernels. By employing the internal recurrence of information inside an image, ZSSR \cite{ZSSR} can use pseudo image pairs to recover LR images with diverse blur kernels. Inspired by CycleGAN, Yuan $et\ al$. \cite{CinCGAN} proposed CinCGAN, which first generates bicubic downsampled LR images from the input and then super-resolves LR images to HR images. However, CinCGAN only considers single bicubic degradation, resulting in poor generalization in complicated real-world SR tasks. By contrast, our method utilizes UBCDTN to simulate diverse real-world degradation types, making it perform proficiently in the real world. 

\section{Overview}
We divide the unsupervised super-resolution problem into two stages, but it is still in an end-to-end fashion. The proposed method comprises two stages: 1) Unsupervised image domain translation between real-world LR images and artificially degraded LR images. 2) Semantic encoder guided generative adversarial super resolution from the generated real-like LR images to final HR images in a supervised manner. To obtain the artificially degraded LR image $I^{LR}_{degraded}$, we perform bicubic downscaling on HR image $I^{HR}$ by image degradation block.

In the first stage, the proposed UBCDTN generates real-like LR images belonging to the real-world domain from the artificially degraded LR images. It aims to transfer the domain of artificially degraded images to the real-world domain, which can ensure a similar real-world LR pattern in the generated real-like LR images. After domain translation performed by UBCDTN, we obtain the real-like LR $\widehat{I}^{LR}_{real}$ treated as the input of the following SESRN. In the second stage, we utilize SESRN to learn the super-resolution mapping from the real-like LR space to the HR space. We further train the SESRN with adversarial loss, content loss, and pixel-wise loss to generate photo-realistic SR images.  

\subsection{Unsupervised Bi-directional Cycle Domain Transfer Network}
The designed UBCDTN is able to translate the source domain $I^{LR}_{degraded}$ to the target domain $I^{LR}_{real}$. It provides an effective bi-directional cycle solution to reduce the domain gap between $I^{LR}_{degraded}$ and $I^{LR}_{real}$. The forward-cycle module contains $G_{A}$ where it aims to learn a mapping $I^{LR}_{degraded} \rightarrow \widehat{I}^{LR}_{real}$. The goal of $G_{B}$ in backward-cycle module is to learn another mapping $I^{LR}_{real} \rightarrow I^{LR}_{syn}$.

As shown in Fig. 1, the forward-cycle module comprises generator $G_{A}$, discriminator $D_{B}$ and Feature Extractor $FE_{A}$. The backward-cycle module consists of $G_{B}$, $D_{A}$, and $FE_{B}$, where these networks have the same architecture as the forward-cycle module's corresponding components, but for different purposes. We utilize U-Net as the basic architecture for $G_{A}$ and $G_{B}$. $D_{A}$ and $D_{B}$ are standard convolutional neural networks, where each is composed of nine convolutional layers followed by BatchNormalization (BN) layers and Leaky ReLU layers. The pre-trained VGG19 is exploited as feature extractor $FE_{A}$ and $FE_{B}$.

The UBCDTN trains two generators simultaneously, where these two generators should be translated in bi-direction and inverted to each other. We involve adversarial learning in both modules, where $G_{A}$ and $G_{B}$ are trained on discriminator $D_{B}$, $D_{A}$ respectively. However, as mentioned before, the SISR task is an ill-posed problem in which there exist many possible SR images reconstructed from one given LR input. Thus, applying adversarial loss alone to train UBCDTN leads to the problem of model collapse, and it cannot produce the desired generated images which have the same characteristics and patterns as target domain images. It is necessary to employ regularization on UBCDTN to improve the quality of translated images. Thus, we propose a cycle-consistency constraint to guarantee the domain correlation between ($\widehat{I}^{LR}_{real}$, $I^{LR}_{real}$) and ($I^{LR}_{syn}$, $I^{LR}_{degraded}$). In addition, in order to avoid color variation, identity loss is applied to preserve color composition between the input image and output image \cite{CycleGAN}. Moreover, we introduce the cycle-perceptual loss as an additional constraint to preserve the sharper edges and finer structures in the reconstructed images. The following sections provide more details of the forward-cycle module and backward-cycle module. 

\subsubsection{Forward-cycle Module}
The forward-cycle module contains a generator $G_{A}$, discriminator $D_{B}$ and feature extractor $FE_{A}$. The goal of $G_{A}$ is to map the artificially degraded LR image to the target domain image, i.e., $\widehat{I}_{real}^{LR} = G_{A}((I^{LR}_{degraded})_{i})$. The concern of the unpaired condition can be eliminated by adding additional constraints. We propose the cycle-consistency to guarantee the relevant content in the generated $\widehat{I}^{LR}_{real}$ can be preserved, i.e., $G_{B}(G_{A}((I^{LR}_{degraded})_{i})) \approx I^{LR}_{degraded}$. In a word, the generated images of $G_{A}$ and $G_{B}$ should be cycle-consistent with each other.  Thus, the forward cycle consistency is established  $I^{LR}_{degraded} \rightarrow G_{A}(I^{LR}_{degraded}) \rightarrow G_{B}(G_{A}(I^{LR}_{degraded})) \approx I^{LR}_{degraded}$. The forward cycle consistency loss $L^{cyc}_{G_{B}}$ can be defined as follows:
{
\begin{equation}
\widehat{I}_{recon}^{LR} = G_{B}(G_{A}((I^{LR}_{degraded})_{i}))
\end{equation}
\begin{equation}
\begin{split}
L^{cyc}_{G_{B}} = \frac{1}{N}\sum_{i}^{N}||(\widehat{I}^{LR}_{recon})_{i} - (I^{LR}_{degraded})_{i} ||_{1}
\end{split}
\end{equation}}where $N$ is the number of training images. As Eqn. 2 shows, with the help of $G_{B}$, $\widehat{I}^{LR}_{recon}$can be identical to the input $I^{LR}_{degraded}$. Through the cycle consistency mechanism, $I^{LR}_{degraded}$ in the source domain can be reconstructed after performing $G_{A}$ and $G_{B}$ on $I^{LR}_{degraded}$ in turn. In addition, we apply adversarial losses to $G_{A}$ and $D_{B}$ such that the distribution of $\widehat{I}^{LR}_{real}$ is  indistinguishable from the real distribution $I^{LR}_{real}$. We apply the Relativistic average GAN (RaGAN) \cite{RaGAN} to train $G_{A}$ and $D_{B}$. $G_{A}$ receives the input $I^{LR}_{degraded}$ and generates $\widehat{I}^{LR}_{real}$, which can fool $D_{B}$. The discriminator $D_{B}$ aims to predict the probability that the provided real image $I^{LR}_{real}$ is more realistic compared to the generated fake image $\widehat{I}_{real}^{LR}$. The input of $D_{B}$ contains real data $I^{LR}_{real}$ and fake data $\widehat{I}_{real}^{LR}$ respectively. It can be summarized as follows:
{
\begin{equation}
D(X_{real}) = \sigma (C(I^{LR}_{real}) - E_{x_{f}}[C(\widehat{I}_{real}^{LR})])
\end{equation}
\begin{equation}
D(X_{fake}) = \sigma (C(\widehat{I}_{real}^{LR}) - E_{x_{r}}[C(I^{LR}_{real})])
\end{equation}}where $\sigma$ denotes sigmoid non-linearity, $C$ denotes the discriminator without the final sigmoid layer, $E_{x_{f}}$, $E_{x_{r}}$ are mathematical means of $X_{fake}$ and $X_{real}$ in the training batch respectively, and $D(X_{real})$, $D({X_{fake}})$ are the predicted probability of $I^{LR}_{real}$, $\widehat{I}_{real}^{LR}$ to be real by discriminator $D_{B}$. An adversarial loss for $G_{A}$, $D_{B}$ can be expressed as $L^{adv}_{G_{A}}$ and $L^{adv}_{D_{B}}$ respectively as follows:
{
\begin{equation}
\begin{split}
L^{adv}_{G_{A}}= & -\mathbb{E}_{I^{LR}_{real}}\sim p_{(I^{LR}_{real})} [ log ( 1- D(X_{real})) ] \\
& -\mathbb{E}_{\widehat{I}_{real}^{LR}}\sim p_{(\widehat{I}_{real}^{LR})}[ log ( D( X_{fake} ) ) ]
\end{split}
\end{equation}}
{
\begin{equation}
\begin{split}
L^{adv}_{D_{B}}= & -\mathbb{E}_{I^{LR}_{real}}\sim p_{(I^{LR}_{real})} [ log ( D(X_{real})) ] \\
& -\mathbb{E}_{\widehat{I}_{real}^{LR}}\sim p_{(\widehat{I}_{real}^{LR})}[ log ( 1-D( X_{fake} ) ) ]
\end{split}
\end{equation}}where $\mathbb{E}_{I^{LR}_{real}}\sim p_{(I^{LR}_{real})}$ and  $\mathbb{E}_{\widehat{I}_{real}^{LR}}\sim p_{(\widehat{I}_{real}^{LR})}$ indicates the distribution of real image $I^{LR}_{real}$, and fake image $\widehat{I}^{LR}_{real}$ respectively. We further introduce forward identity loss $L^{idt}_{degraded}$ to maintain color composition between $\widehat{I}^{LR}_{idt}$ and $I^{LR}_{degraded}$. The $\widehat{I}^{LR}_{idt}$ can be generated by $G_{B}$, as Eqn. 8 shows. The $L^{idt}_{degraded}$ is expressed as Eqn. 9:
{
\begin{equation}
\widehat{I}_{idt}^{LR} = G_{B}((I^{LR}_{degraded})_{i})
\end{equation}
\begin{equation}
L^{idt}_{degraded} = \frac{1}{N}\sum_{i}^{N}||(\widehat{I}^{LR}_{idt})_{i} - (I^{LR}_{degraded})_{i} ||_{1}
\end{equation}}Moreover, in order to minimize the perceptual divergence between $\widehat{I}^{LR}_{recon}$ and $I^{LR}_{degraded}$, we utilize $FE_{A}$ to extract VGG feature maps for cycle-perceptual loss estimation. It can be defined as follows:
{
\begin{equation}
\begin{split}
L^{percep}_{FE_{A}}= \frac{1}{N}\sum_{i}^{N}|| FE_{q,r}((\widehat{I}^{LR}_{recon})_{i}) - FE_{q,r}((I^{LR}_{degraded})_{i})||_{2}
\end{split}
\end{equation}}where $FE_{q,r}(\cdot)$ indicates the extracted feature maps of $q$-th convolution layer (after activation layer) before $r$-th maxpooling layer. The total objective loss for forward cycle module $L^{Forward}_{total}$ is a weighted sum of the four loss functions:

\begin{equation}
\begin{split}
L^{Forward}_{total} = \omega_{1}L^{adv}_{G_{A}}+\omega_{2}L^{cyc}_{G_{B}} +\omega_{3}L^{idt}_{degraded\_LR}+\omega_{4}L^{percep}_{FE_{A}}
\end{split}
\end{equation}where the hyper-parameters $\omega_{1}$, $\omega_{2}$, $\omega_{3}$, $\omega_{4}$ are trade-off factors for each loss. We empirically set $\omega_{1}$, $\omega_{2}$, $\omega_{3}$, and $\omega_{4}$ to 1, 10, 1 and 1 respectively. It is noticeable that the loss with high weights indicates a significant proportion of the training process. 

\subsubsection{Backward-cycle Module}
To transfer the image from the target domain to the source domain, i.e., $I^{LR}_{real} \rightarrow I^{LR}_{syn}$, we specifically construct a backward-cycle module in which the generator $G_{B}$ is able to learn the mapping $G_{B}(I^{LR}_{real}) \approx I^{LR}_{syn}$. The learned mapping is capable of re-expressing the target domain image $I^{LR}_{real}$ by the source domain image $I^{LR}_{syn}$ implicitly. 
The backward-cycle module consists of a generator $G_{B}$, discriminator $D_{A}$ and feature extractor $FE_{B}$. There are several constraints required in the backward-cycle module. First, we symmetrically design the backward cycle consistency constraint, i.e., $G_{A}(G_{B}((I^{LR}_{real})_{i})) \approx I^{LR}_{real}$, which further forces the contents of generated images to be relevant to the input ones. Thus, the backward cycle is composed of $I^{LR}_{real} \rightarrow G_{B}(I^{LR}_{real}) \rightarrow G_{A}(G_{B}(I^{LR}_{real})) \approx I^{LR}_{real}$. The backward cycle consistency loss can be defined as follows:
{
\begin{equation}
I_{recon}^{LR} = G_{A}(G_{B}((I^{LR}_{real})_{i}))
\end{equation}
\begin{equation}
L^{cyc}_{G_{A}}= \frac{1}{N}\sum_{i}^{N}||(I^{LR}_{recon})_{i}) - (I^{LR}_{real})_{i} ||_{1}
\end{equation}}As Eqn. 12 shows, because of the formulated backward cycle scheme, $G_{B}$ and $G_{A}$ are capable of inverting the reconstructed images $I^{LR}_{recon}$ back to the $I^{LR}_{real}$. The Eqn. 13 is able to minimize the correlation discrepancy between $I^{LR}_{recon}$ and $I^{LR}_{real}$ such that the consistent characteristics can be preserved. Similarly, as the forward cycle module, we also adopt adversarial learning in the backward cycle module, where the generator $G_{B}$ and discriminator $D_{A}$ are optimized by RaGAN. The adversarial loss $L^{adv}_{G_{B}}$ and  $L^{adv}_{D_{A}}$ are defined as:
{
\begin{equation}
D(X_{real}) = \sigma (C(I^{LR}_{degraded}) - E_{x_{f}}[C(I_{syn}^{LR})])
\end{equation}}
{
\begin{equation}
D(X_{fake}) = \sigma (C(I_{syn}^{LR}) - E_{x_{r}}[C(I^{LR}_{degraded})])
\end{equation}
{
\begin{equation}
\begin{split}
L^{adv}_{G_{B}}= & -\mathbb{E}_{I^{LR}_{degraded}}\sim p_{(I^{LR}_{degraded})} [ log ( 1- D(X_{real})) ] \\
& -\mathbb{E}_{I_{syn}^{LR}}\sim p_{(I_{syn}^{LR})}[ log ( D( X_{fake} ) ) ]
\end{split}
\end{equation}}
{
\begin{equation}
\begin{split}
L^{adv}_{D_{A}}= & -\mathbb{E}_{I^{LR}_{degraded}}\sim p_{(I^{LR}_{degraded})} [ log ( D(X_{real})) ] \\
& -\mathbb{E}_{I_{syn}^{LR}}\sim p_{(I_{syn}^{LR})}[ log ( 1-D( X_{fake} ) ) ]
\end{split}
\end{equation}}According to the adversarial loss, $G_{B}$ can generate desired $I^{LR}_{syn}$ under the supervision of $D_{A}$. In order to avoid color variation between $I^{LR}_{idt}$ and $I^{LR}_{real}$, we further define backward identity loss $L^{idt}_{real}$. The $I^{LR}_{idt}$ is generated by $G_{A}$, as Eqn. 18 shows and the $L^{idt}_{real\_LR}$ is defined as Eqn. 19:
{
\begin{equation}
I_{idt}^{LR} = G_{A}((I^{LR}_{real})_{i})
\end{equation}
\begin{equation}
L^{idt}_{real\_LR}= \frac{1}{N}\sum_{i}^{N}||(I^{LR}_{idt})_{i} - (I^{LR}_{real})_{i} ||_{1}
\end{equation}}Moreover, the backward cycle-perceptual loss $L^{percep}_{FE_{B}}$ is calculated to recover visual pleasing details. We utilize $FE_{B}$ to measure the Euclidean distance between feature maps of $I^{LR}_{recon}$ and $I^{LR}_{real}$. It can be defined as follows:
{
\begin{equation}
\begin{split}
L^{percep}_{FE_{B}}= \frac{1}{N}\sum_{i}^{N}|| FE_{q,r}((I^{LR}_{recon})_{i}) - FE_{q,r}((I_{real}^{LR})_{i})||_{2}
\end{split}
\end{equation}}In the end, the total optimization loss $L^{Backward}_{total}$ for the backward cycle module can be formulated as:
{
\begin{equation}
\begin{split}
L^{Backward}_{total}= \lambda _{1}L^{adv}_{G_{B}}+\lambda_{2}L^{cyc}_{G_{A}} +\lambda_{3}L^{idt}_{real\_LR}+\lambda_{4}L^{percep}_{FE_{B}}
\end{split}
\end{equation}}where the hyper-parameters $\lambda_{1}$, $\lambda_{2}$, $\lambda_{3}$, $\lambda_{4}$ are the set of corresponding weights for $L^{adv}_{G_{B}}$, $L^{cyc}_{G_{A}}$, $L^{idt}_{real\_LR}$ and $L^{percep}_{FE_{B}}$. The $\lambda_{1}$, $\lambda_{2}$, $\lambda_{3}$, and $\lambda_{4}$ are empirically set to 1, 10, 1 and 1 respectively.

\subsubsection{Total Unsupervised Bi-directional Cycle Domain Transfer Network Loss}
The full optimization objective loss of UBCDTN $L^{UBCDTN}_{total}$, which is an addition of forward cycle module loss $L^{Forward}_{total}$ and backward cycle module loss $L^{Backward}_{total}$, can be represented as follows:
{
\begin{equation}
L^{UBCDTN}_{total} = L^{Forward}_{total} + L^{Backward}_{total}
\end{equation}}

\begin{figure*}[t!]
\centerline{\includegraphics[width =\textwidth]{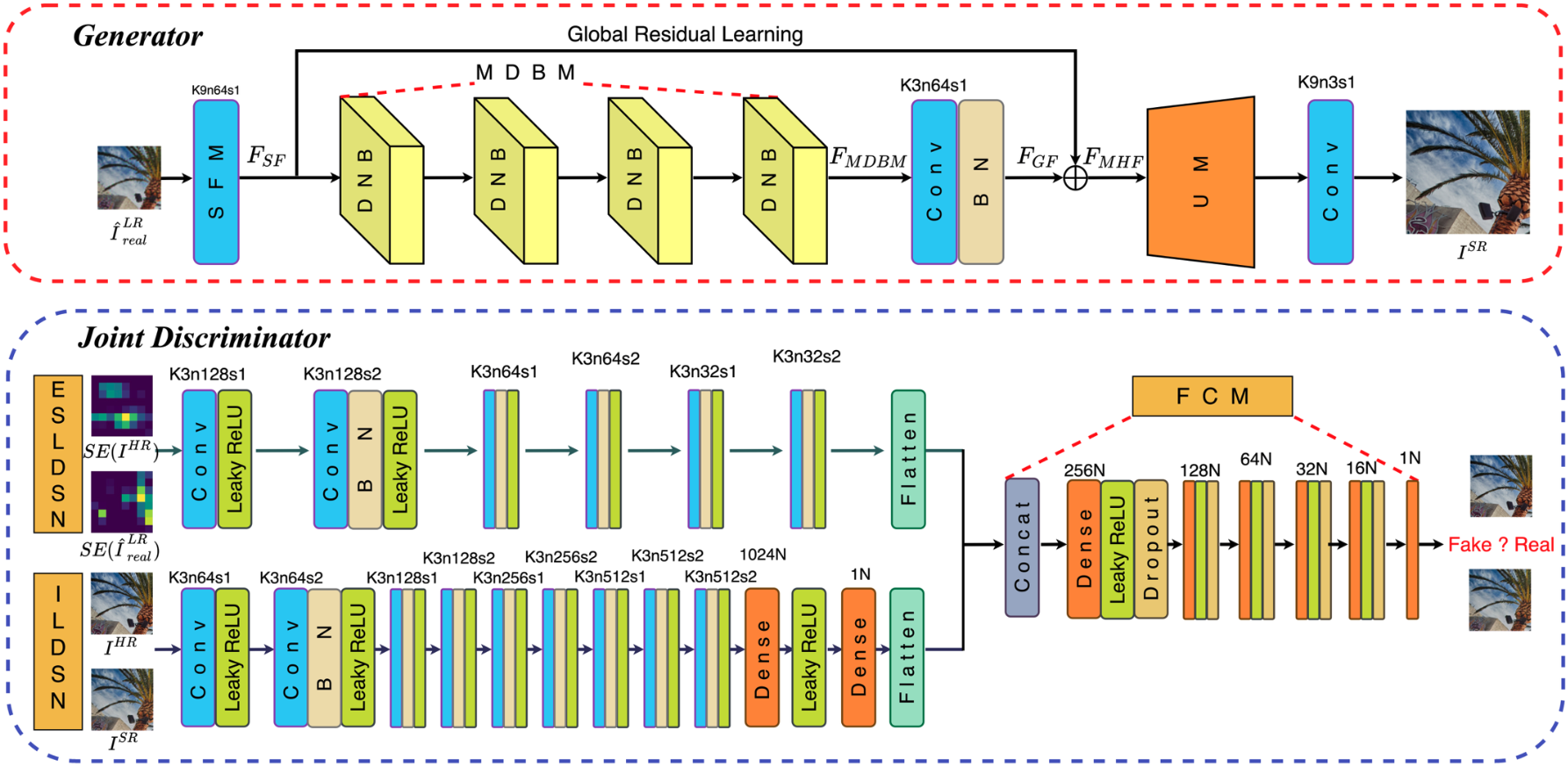}}
\caption{\textbf{Red dotted rectangle}: The architecture of the Generator. \textbf{Blue dotted rectangle}: The architecture of the Joint Discriminator. $F_{SF}$ denotes shallow features, $F_{MDBM}$ denotes the outputs of MDBM, $F_{GF}$ represents global features, and $F_{MHF}$ represents multiple hierarchical features. K, n and s are the kernel size, number of filters, and strides respectively. N is the number of neurons in the dense layer.}
\label{fig}
\end{figure*}

\subsection{Semantic Encoder Guided Super Resolution Network}
In this section, we present the proposed SESRN, which aims to generate the desired $I^{SR}$ from the $\widehat{I}^{LR}_{real}$ produced by UBCDTN. First, we describe four main components of SESRN: $G_{SR}$, $SE$, $D_{SR}$, $CE$. Second, we describe the optimization loss function for SESRN. The architecture of SESRN is illustrated in Fig. 2. In addition, the structure of the joint discriminator and generator proposed in GAFH-RIDN \cite{xiang2021end-to-end} are shown in Fig. 3. 

\subsubsection{Generator}
We utilize the Dense Nested Blocks (DNBs) and Residual in Internal Dense Blocks (RIDBs) that we proposed in GAFH-RIDN \cite{xiang2021end-to-end} as basic units to construct $G_{SR}$. The goal of $G_{SR}$ is to super-resolve to the SR image $I^{SR}$ from the real-like LR image $\widehat{I}_{real}^{LR}$. The overall super-resolution process is formulated as:
{
\begin{equation}
I^{SR}=G_{SR}(\widehat{I}_{real}^{LR})
\end{equation}}As shown at the top of Fig. 3, the generator $G_{SR}$ contains three stages: Shallow Feature Module (SFM), Multi-level Dense Block Module (MDBM), and Upsampling Module (UM). Specifically, the MDBM is built up by multiple DNBs formed by several RIDBs. Each DNB includes 3 RIDBs cascaded by residual connections and one scaling layer. The designed architecture guarantees the feature maps of each layer are propagated into all succeeding layers, promoting an effective way to extract hierarchical features and alleviating the gradient vanishing problem. It is emphasized that Local Residual Learning (LRL) is introduced to take effective use of the local residual features extracted by RIDBs. In addition, in order to help the generator fully take advantage of hierarchical features, we design the Global Residual Learning (GRL) to fuse the shallow features $F_{SF}$ and global features $F_{GF}$. Overall, benefiting from the proposed architecture \cite{xiang2021end-to-end}, $G_{SR}$ is capable of exploiting abundant hierarchical features and super-resolving from the LR space to the HR space.

\subsubsection{Semantic Encoder}
The proposed semantic encoder is supposed to extract embedded semantics (as shown in Fig. 2), which is used to project visual information (HR, LR) back to the latent space. The motivation is that the GAN-based SR models \cite{SRGAN,ESRGAN_Wang,URDGN} only exploit visual information during the discriminative procedure, ignoring the semantic information reflected by latent representation. Therefore, the proposed semantic encoder will complement the missing critical property. Previous GAN's work \cite{BiGAN,ALI} has proved that the semantic representation is beneficial to the discriminator.

Based on this observation, the proposed semantic encoder is designed to inversely map the image to the embedded semantics. Significantly, the most important advantage of the semantic encoder is that it is able to guide the discriminative process since the embedded semantics obtained from the semantic encoder can reflect semantic attributes, such as the image features (color, texture, and shape) and the spatial relationship between various components of the images. It can be emphasized that the embedded semantics is fed into a joint discriminator along with HR and LR images, leading to optimizing the adversarial process. Thanks to this property, the semantic encoder can guide the discriminative adversarial learning of the joint discriminator, thereby enhancing its discriminative ability.

In this context, we utilize the pre-trained VGG16 \cite{VGG16} networks as the semantic encoder. For the purpose of satisfying the different dimensions of HR and LR images, we adopt two side-by-side semantic encoders, which have the same structure but different input dimensions, to obtain embedded semantics from different convolutional layers respectively. 

\subsubsection{Joint Discriminator}
As shown in Fig. 2, the proposed joint discriminator takes the tuple incorporating both visual information and embedded semantics as the input, where Embedded Semantics-Level Discriminative Sub-Net (ESLDSN) aims to identify the embedded semantics whether it comes from the HR images while Image-Level Discriminative Sub-Net (ILDSN) distinguishes whether the input image is from the HR image dataset or the generator. Next, through the operation of the Fully Connected Module (FCM) on a concatenated vector, the final probability is predicted.

Thanks to this property, the joint discriminator is capable of learning the joint probability distribution of image data ($I^{HR},I^{SR}$) and embedded semantics ($SE(I^{HR}),SE(\widehat{I}^{LR}_{real})$). In order to satisfy this structure, we design two sets of paths entering into the joint discriminator. The set of red paths in Fig. 2 represents a real tuple which consists of the real HR image $I^{HR}$ from the dataset and its embedded semantics $SE(I^{HR})$. The set of paths in blue indicates the fake tuple constructed by SR image $I^{SR}$ generated from the generator and $SE(\widehat{I}^{LR}_{real})$ produced by semantic encoder through real-like LR image. Therefore, the joint discriminator has the ability to measure the difference between real tuple $(I^{HR},SE(I^{HR}))$ and the fake tuple $(I^{SR},SE(\widehat{I}^{LR}_{real}))$. 

In addition, we adopt the Relativistic average Least Squares GAN (RaLSGAN) \cite{RaGAN} loss for the joint discriminator by applying the RaD to the least squares loss function \cite{LSGAN}. The real tuple is denoted as $X_{real}=(I^{HR},SE(I^{HR}))$, and the fake tuple is expressed as $X_{fake} = (I^{SR},SE(\widehat{I}^{LR}_{real}))$. In the training procedure, the joint discriminator receives both $X_{real}$ and $X_{fake}$ as the input. It can be expressed as follows:
{
\begin{equation}
\tilde{C}(X_{real}) = (C(X_{real}) - E_{x_{f}}[C(X_{fake})])
\end{equation}
\begin{equation}
\tilde{C}(X_{fake}) = (C(X_{fake}) - E_{x_{r}}[C(X_{real})])
\end{equation}}where $\tilde C(\cdot)$ denotes the probability predicted by the joint discriminator. Moreover, the least squares loss is applied to evaluate the distance between HR and SR images. Thus, we define the optimization loss $L_{D_{SR}}^{RaLS}$ and $L_{G_{SR}}^{RaLS}$ for joint discriminator and generator respectively:
{
\begin{equation}
\begin{split}
L_{D_{SR}}^{RaLS}= & 
\mathbb{E}_{I^{HR}\sim p_{(I^{HR})}}[( \tilde{C}( X_{real})-1)^{2}] \\
& +\mathbb{E}_{I^{SR}\sim p_{(I^{SR})}}[( \tilde{C}( X_{fake})+1)^{2}]
\end{split}
\end{equation}
\begin{equation}
\begin{split}
L_{G_{SR}}^{RaLS}=& \mathbb{E}_{I^{SR}\sim p_{(I^{SR})}}[( \tilde{C}( X_{fake})-1)^{2}] \\ &+\mathbb{E}_{I^{HR}\sim p_{(I^{HR})}}[( \tilde{C}( X_{real})+1)^{2}]
\end{split}
\end{equation}}By taking advantage of least squares loss and relativism in RaLS, SESRN is capable of enhancing model stability and producing visually pleasant SR images.

\subsubsection{Content Extractor}
In SESRN, we further leverage the pre-trained VGG19 network as content extractor $\phi$ to obtain feature representations, where they are used to formulate the content loss $L_{content}$. Specifically, we calculate the $L_{content}$ based on the Euclidean distance between two feature representations of SR images and HR images. We extract feature representations from the `Conv3\_3' layer in the content extractor, which are the low-level features before the activation layer. With this loss term, the SESRN is encouraged to reconstruct finer high-frequency details and improve perceptual quality.

\subsubsection{Loss Function}
We introduce content loss $L_{content}$ to constrain SR images to be faithful to human visual perception. Besides, we also involve pixel-wise loss $L_{pixel}$ to optimize our method. Furthermore, adversarial losses $L_{G_{SR}}^{RaLS}$ and $L_{D_{SR}}^{RaLS}$ are applied to $G_{SR}$ and $D_{SR}$ respectively, which allows the generator to produce the SR image consistent with the distribution of the HR image.

{Content Loss}:
$L_{content}$ is able to improve the perceptual similarity between SR and HR images. It is formulated as:
{
\begin{equation}
L_{content}=|| \phi_{i,j}(I^{HR})-\phi_{i,j}( I^{SR})||^{2}_{2}
\end{equation}}where $\phi_{i,j}(\cdot)$ represents the feature representations obtained from $j$-th convolution layer before $i$-th maxpooling layer in the fixed content extractor.

{Pixel-wise Loss}: The pixel-wise loss is widely applied to enforce the intensity similarity between the SR and HR images. It is calculated as:
{
\begin{equation}
L_{pixel} = \frac{1}{N}\sum_{i}^{N}||(I^{SR})_{i} - (I^{HR})_{i} ||_{2}
\end{equation}}We use $L_{pixel}$ to minimize the distance between $I^{SR}$ and $I^{HR}$.

{Total Loss}: Finally, we obtain the total loss function $L_{total}^{SESRN}$ for SESRN. It is the weighted sum of the three above-discussed losses. The formula is described as follows:
{
\begin{equation}
L^{SESRN}_{total} = \lambda _{con}L_{content} + \lambda_{adv}L_{G}^{RaLS}+ \lambda_{pixel}L_{pixel}
\end{equation}}where $\lambda_{con}$, $\lambda_{adv}$ and $\lambda_{pixel}$ are empirically set to 1, $10^{-3}$ and 1 respectively.

\subsection{Full objective loss for UBCDTL-GAN}
The full objective loss for the UBCDTL-GAN, which is the combination of $L^{UBCDTN}_{total}$ and $L^{SESRN}_{total}$, can be defined as:
{
\begin{equation}
L^{UBCDTL-GAN}_{total} = L^{UBCDTN}_{total} + L^{SESRN}_{total}
\end{equation}}The complete objective loss $L^{UBCDTL-GAN}_{total}$ encourages the proposed method UBCDTL-GAN can solve the unpaired real-world image super-resolution problem. 

\section{Experiments}
In this section, we first present the datasets and the experimental details. Then, we compare our method with state-of-the-art SISR methods. The qualitative comparison can be seen in Fig. 4, and the quantitative comparisons are provided in Table I. It is emphasized that all the quantitative results are cited from their official published papers. Moreover, as for qualitative comparison, we directly download the released code and pre-trained models of the compared methods. Then, we carefully re-implement referenced methods on the same validation dataset to obtain effective results.

\begin{figure*}[!t]
\centerline{\includegraphics[width=\textwidth]{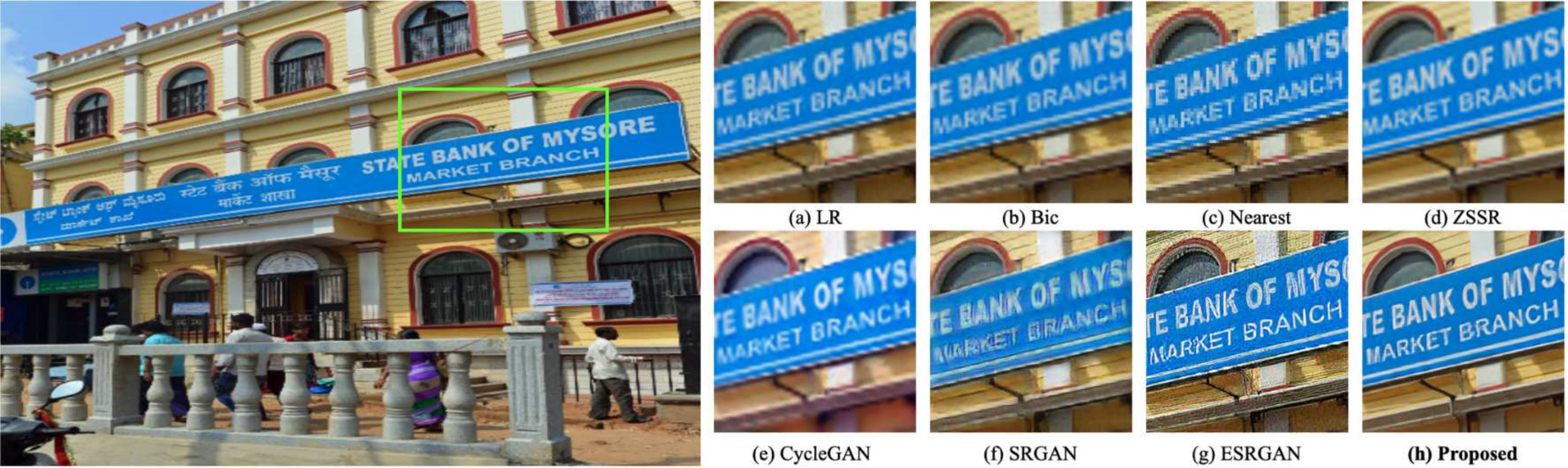}}
\centerline{\includegraphics[width =\textwidth]{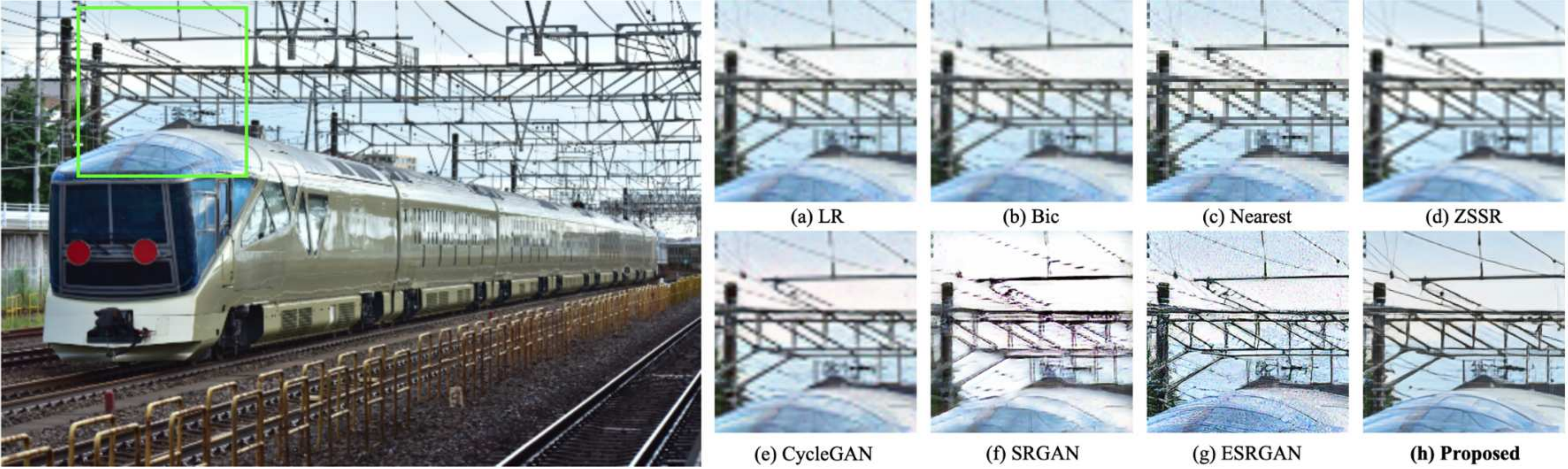}}
\centerline{\includegraphics[width =\textwidth]{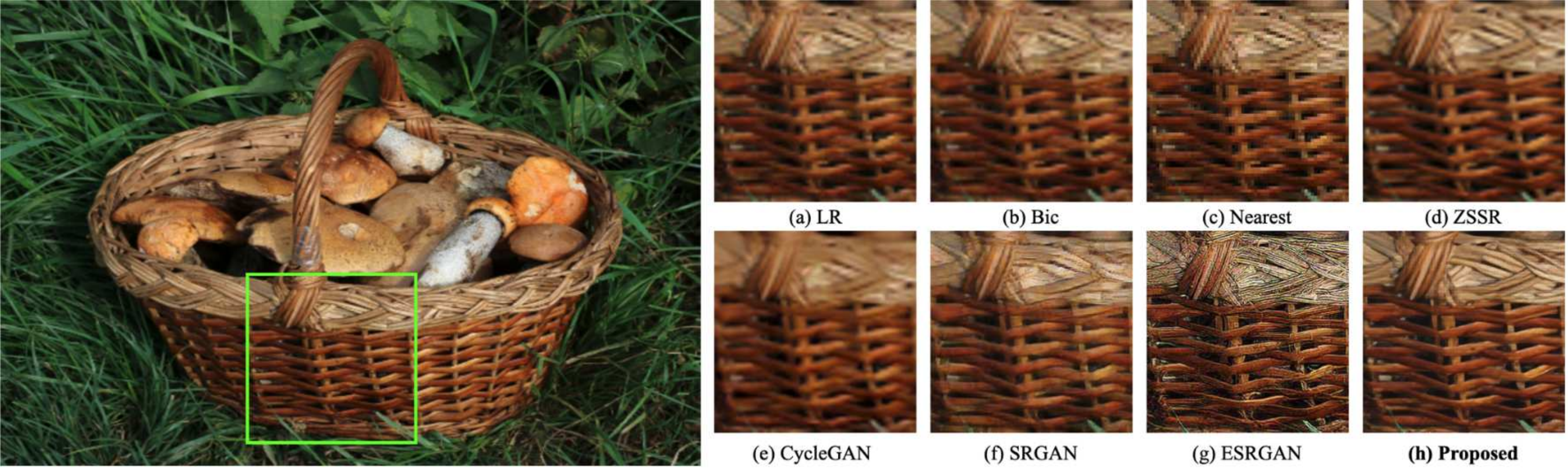}}
\caption{Qualitative comparison of visual results with state-of-the-art methods on NTIRE 2020 Real-World Track 1 images. Our method produces photo-realistic results.}
\label{fig}
\end{figure*}

\subsection{Training Data}
At the training stage, we conduct experiments on the DF2K dataset \cite{DIV2K, EDSR}, which is a merge of the DIV2K and Flickr2K datasets, containing a total of 3450 images. Specifically, as for the LR images, we use the real-world LR images from DIV2K NTIRE 2017 unknown degradation 4× dataset, and Flickr2K LR images collected from NTIRE 2020 Real-World Track 1 training source dataset, where all the LR images are corrupted with unknown degradation, and downsampled 4× by an unpredicted operator to satisfy real-world conditions, resulting in sensor noise, compression artifacts, etc. Since the goal of our method is to solve an unsupervised super-resolution problem without LR-HR paired images, we select the first 1725 images (number: 1-1725) from the DF2K HR dataset as our HR training dataset, and the LR training dataset is formed by the other 1725 images (number: 1726-3450) obtained from DF2K real-world LR dataset. Overall, our method is trained on such an unpaired real-world LR-HR dataset.

To evaluate the proposed method on real-world data, at the testing stage, we use the validation dataset from the NTIRE 2020 Real-World SR challenge Track 1. This dataset contains 100 testing LR images (scaling factor: 4×), where all LR images are processed with unknown degradation operations to simulate the realistic artifacts and natural characteristics. In order to compare the qualitative and quantitative results fairly, we use the same validation dataset for all experiments. 

\subsection{Training Setups}
The training procedure is divided into three steps. Instead of randomly initializing model weights, we pre-train the UBCDTN and SESRN in the first and second steps, and then we jointly train the whole method in an end-to-end manner. First, we train the UBCDTN with unpaired artificially degraded images $I^{LR}_{degraded}$ and real-world images $I^{LR}_{real}$, which aims to transfer the LR image from the artificially degraded LR domain to the real-world LR domain. Second, we pre-train the SESRN using the approximated real-like images $\widehat{I}^{LR}_{real}$ and their HR version $I^{HR}$ to generate realistic super-resolved images $I^{SR}$. As for pre-trained UBCDTN and SESRN, we use the same optimization strategy where the Adam optimizer \cite{Adam} is applied to train both networks by setting $\beta_{1}=0.9$, $\beta_{2}=0.999$. The learning rate is initialized as $10^{-4}$ and the minibatch size is set as 8. We train the UBCDTN and SESRN 50K epochs separately. After the pre-training process, both UBCDTN and SESRN are able to acquire exceptional initialization weights, which are beneficial to training stability and fast training speed. In the third step, we jointly train two networks together. The proposed UBCDTL-GAN takes an unpaired HR image $I^{HR}$ and $I^{LR}_{real}$ as the input to generate the final result $I^{SR}$. The UBCDTL-GAN applies the Adam optimizer with an initial learning rate = $10^{-4}$, $\beta_{1}=0.9$ and $\beta_{2}=0.999$ to the whole training process. The model is trained for 100K epochs with the minibatch size 8.

\begin{table*}[t!]
\caption{Quantitative comparison on NTIRE 2020 Real World Super-Resolution Challenge Track 1 validation dataset of the proposed method against state-of-the-art methods, in terms of average PSNR (dB) and SSIM for upscale factor 4×. The bold results indicate the best performance.}
\small
\centering
\setlength{\tabcolsep}{13.5mm}
\begin{tabular}{lllllcllllllll}
\hline
\hline
\multicolumn{4}{l}{Methods} &  & \multicolumn{9}{c}{NTIRE\_2020\_T1} \\ 
\multicolumn{4}{l}{} &  & \multicolumn{4}{c}{PSNR} &  & \multicolumn{4}{l}{SSIM} \\ \hline
\multicolumn{4}{l}{Bicubic} &  & \multicolumn{4}{c}{23.87} &  & \multicolumn{4}{l}{0.644} \\
\multicolumn{4}{l}{Nearest Neighbor} &  & \multicolumn{4}{c}{23.39} &  & \multicolumn{4}{l}{0.580} \\
\multicolumn{4}{l}{EDSR \cite{EDSR}} &  & \multicolumn{4}{c}{25.36} &  & \multicolumn{4}{l}{0.640} \\
\multicolumn{4}{l}{ESRGAN \cite{ESRGAN_Wang}} &  & \multicolumn{4}{c}{19.04} &  & \multicolumn{4}{l}{0.242} \\
\multicolumn{4}{l}{SRGAN \cite{SRGAN}} &  & \multicolumn{4}{c}{20.78} &  & \multicolumn{4}{l}{0.525} \\
\multicolumn{4}{l}{SRFBN \cite{SRFBN}} &  & \multicolumn{4}{c}{25.37} &  & \multicolumn{4}{l}{0.642} \\
\multicolumn{4}{l}{MsDNN \cite{MSDNN}} &  & \multicolumn{4}{c}{{25.08}} &  & \multicolumn{4}{l}{0.708} \\
\multicolumn{4}{l}{RCAN \cite{RCNN}} &  & \multicolumn{4}{c}{25.31} &  & \multicolumn{4}{l}{0.640} \\
\multicolumn{4}{l}{USISResNet \cite{USISResNet}} &  & \multicolumn{4}{c}{21.71} &  & \multicolumn{4}{l}{0.589} \\
\multicolumn{4}{l}{CycleGAN \cite{CycleGAN}} &  & \multicolumn{4}{c}{25.01} &  & \multicolumn{4}{l}{0.618} \\
\multicolumn{4}{l}{ZSSR \cite{ZSSR}} &  & \multicolumn{4}{c}{24.87} &  & \multicolumn{4}{l}{0.600} \\
\multicolumn{4}{l}{\textbf{Proposed}} &  & \multicolumn{4}{c}{\textbf{26.83}} &  & \multicolumn{4}{l}{\textbf{0.789}} \\ \hline \hline
\end{tabular}
\end{table*}

\subsection{Quantitative Comparison}
The quantitative results presented in Table I demonstrate that our method has promising superiority, achieving the highest 26.83dB/0.789 in terms of PSNR/SSIM values. The results obtained from SRFBN \cite{SRFBN} place the second-best with 25.37dB/0.642. Our method improves the PSNR/SSIM values by 1.46dB/0.147 over their method. EDSR \cite{EDSR} and SRFBN \cite{SRFBN} cannot achieve desired numerical results since they are PSNR-oriented methods and are merely trained on simple degradation images. Moreover, we found that the performances of ESRGAN \cite{ESRGAN_Wang} and SRGAN \cite{SRGAN} are lower than most of the methods in Table I. Besides, from Fig. 5, the visual results of ESRGAN and SRGAN show over-smoothed textures and unrealistic artifacts. Interestingly, we found such a phenomenon has also been presented in \cite{CinCGAN,ZSSR,USISResNet}. The underlying reason is that these methods only train on the simple and clean pre-defined LR images, ignoring the difference of domain distribution between the real-world LR domain and the pre-defined LR domain. Benefiting from the proposed UBCDTN and SESRN, our method has the ability to solve the problem of domain distribution shift and boost quantitative performances when dealing with real-world SR tasks. 

\subsection{Qualitative Comparison}
The visual comparisons are provided in Fig. 4. We compare with various SR methods such as Bicubic, Nearest Neighbor, SRGAN \cite{SRGAN}, ESRGAN \cite{ESRGAN_Wang}, CycleGAN \cite{CycleGAN}, and ZSSR \cite{ZSSR}. For the Bicubic and Nearest Neighbor methods, it is obvious that the results lack high-frequency contents, producing overly smooth edges and coarse textures. Regarding the SRGAN, it fails to alleviate the blurring details on the lines and edges of the SR results. The results of ESRGAN suffer from apparently broken artifacts and dramatic degradation problems, which are unfaithful to human perception. As for the unsupervised method CycleGAN and ZSSR, the SR results are improved, but there are still far away from the ground truth. Although the SR images of CycleGAN present better shapes than the previously compared methods, the results still remain unnatural edges and distortions, leading to poor visual effects. Besides, the blind method ZSSR was also evaluated, but it fails to reduce visible corruptions in some degree since there are still over-smoothed textures and noise-like characteristics existing in the images. 

Compared with the aforementioned methods, the SR results of our method superiorly outperform all other methods, super-resolving visually pleasant SR images with sharper edges and finer textures. Our SR results are more realistic than SRGAN and ESRGAN since these two methods are only trained on simple degradation data (e.g., bicubiced LR images) without introducing any complicated noise and artifacts from the real-world images while our method trains on approximated real-like LR images consisting of similar characteristics as real-world LR images. The unsupervised method CycleGAN is less effective in super-resolve unclear LR images. Although it involves the cycle translation model, it lacks a powerful super-resolution network as the one (SESRN) in our method. Besides, the other unsupervised method ZSSR also fails to achieve the expected results, since it does not take into account the domain gap between noise-free LR images and real-world images. In contrast, benefiting from the domain transfer network (UBCDTN), our method is able to successfully eliminate the domain gap and produce real-like LR images comprising real-world patterns. Overall, The SR results verify the powerful unsupervised learning strategy used in the proposed method for super-resolving photo-realistic SR images.

\begin{table*}[t]
\caption{The compared variants of the proposed method in the ablation study and the descriptions of the proposed components. The tick indicates that this variant includes this component.}
\small
\setlength{\tabcolsep}{3.8mm}
\begin{tabular}{lllllllll}
\hline
\hline
Variants & \multicolumn{1}{c}{Methods} & SESRN & \multicolumn{1}{c}{$G_{A}$} & $G_{B}$ & $D_{B}$ & $D_{A}$ & $FE_{B}$ & $FE_{A}$ \\ \hline
VariantA & SESRN & $\surd$ &  &  &  &  &  &  \\
VariantB & SESRN+$G_{A}$+$D_{B}$ & $\surd$ & $\surd$ &  & $\surd$ &  &  &  \\
VariantC & {SESRN+$G_{A}$+$FE_{B}$+$G_{B}$+$FE_{A}$} & $\surd$ & {$\surd$} & $\surd$ &  &  & $\surd$ & $\surd$ \\
VariantD & SESRN+$G_{A}$+$D_{B}$+$G_{B}$+$D_{A}$ & $\surd$ & $\surd$ & $\surd$ & $\surd$ & $\surd$ &  &  \\
\textbf{VariantE} & SESRN+$G_{A}$+$D_{B}$+$G_{B}$+$D_{A}$+$FE_{B}$+$FE_{A}$ & $\surd$ & $\surd$ & $\surd$ & $\surd$ & $\surd$ & $\surd$ & $\surd$ \\ \hline \hline
\end{tabular}
\end{table*}

\section{Ablation Study}
In this section, we conduct the ablation study to further investigate the components of the proposed method and demonstrate the advantages of UBCDTL-GAN. The list of compared variants of our method is presented in Table II. We provide visual SR results of different variants in Fig. 5. The quantitative comparison of several variants is presented in Table III.

\subsection{Description of Different Variants of the Proposed Method}
In ablation studies, we design several variants which consist of different proposed components. We adopt the SESRN as the baseline variant in the following experiments and pay more attention to investigating the elements used in the UBCDTN. To comply with the single variable principle, we gradually add one of the components to the baseline variant. We describe the details of designed variants, all of which are specified as follows: 

\textbf{1) VariantA:} The VariantA is designed as the baseline variant, which only contains SESRN. As shown in Table II, VariantA can be considered as removing all UBCDTN components from the ultimate proposed method. In the following variants, we successively add each of the components to VariantA.

\textbf{2) VariantB:} In VariantB, we introduce $G_{A}$ and $D_{B}$ while $G_{B}$, $D_{A}$ and both $FE_{A}$, $FE_{B}$ are removed. Because $G_{A}$ and $D_{B}$ are essential components of the forward cycle module in UBCDTN, VariantB can be considered as composed of the forward cycle module of UBCDTN and baseline model, removing the backward cycle module.

\textbf{3) VariantC:} Besides the baseline model, it consists of two generators $G_{A}$, $G_{B}$ and two feature extractors $FE_{A}$, $FE_{B}$ of UBCDTN, eliminating discriminators $D_{A}$ and $D_B$ involved in UBCDTN. 

\textbf{4) VariantD:} It is constructed by the four components $G_{A}$, $G_{B}$, $D_{B}$ and $D_{A}$ of UBCDTN, while it removes feature extractors $FE_{A}$ and $FE_{B}$ of UBCDTN.

\textbf{5) VariantE (Proposed):} The VariantE represents the ultimate proposed method which comprises a baseline model and all components of UBCDTN.

\begin{figure*}[!t]
 \centerline{\includegraphics[width =\textwidth]{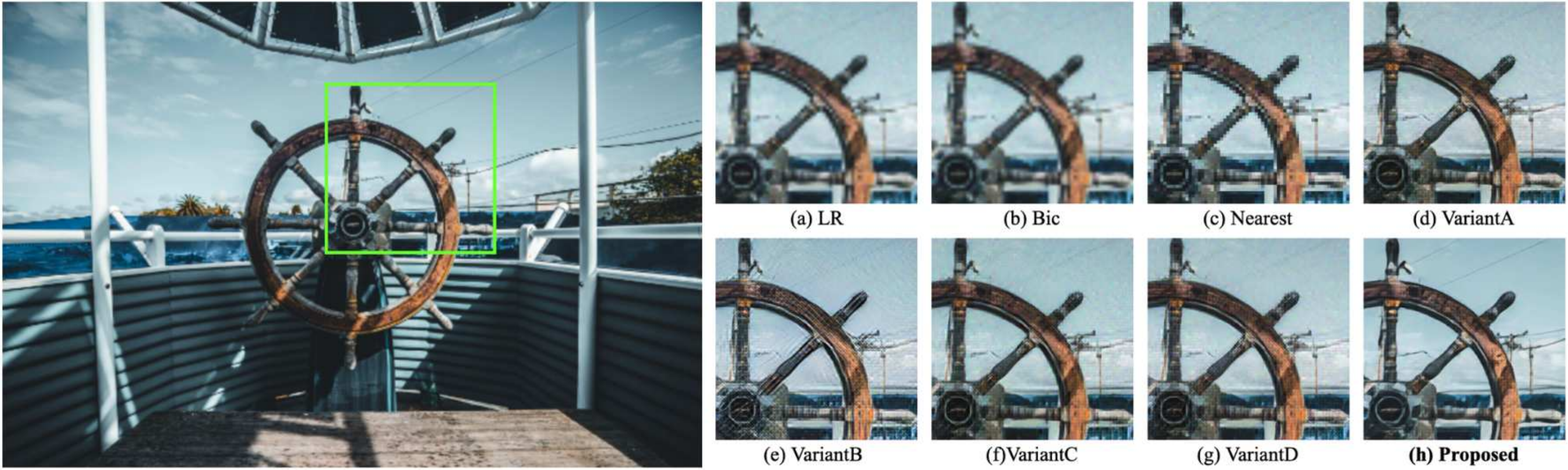}}
 \centerline{\includegraphics[width =\textwidth]{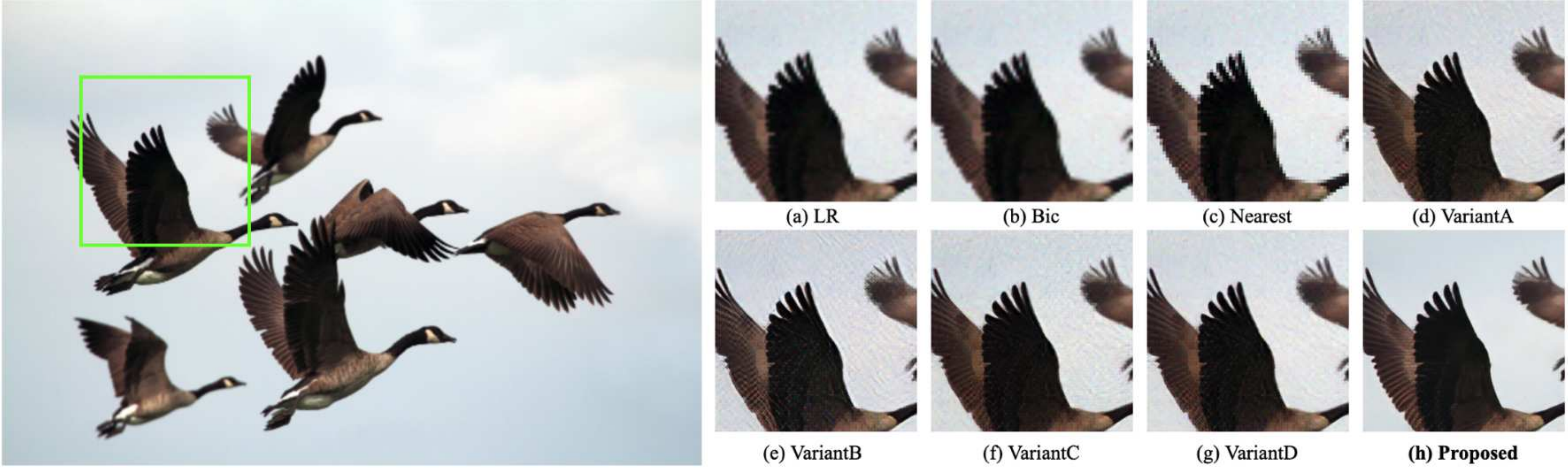}}
 \caption{Qualitative comparisons of different variants in our ablation study. The visual results on images "0896", "0824" from NTIRE 2020 Track 1 testing dataset with scale factor 4×. The best results are highlighted }
\label{fig}
\end{figure*}

\begin{table*}[t!]
\caption{Quantitative results of ablation study with different variants on NTIRE 2020 validation T1 dataset, in terms of average PSNR (dB) and SSIM for upscale factor 4×. The bold results indicate the best performance.}
\small
\centering
\setlength{\tabcolsep}{22.5mm}
\begin{tabular}{llllclllclll}
\hline
\hline
\multicolumn{4}{l}{Variants} & \multicolumn{8}{c}{NTIRE\_2020\_T1} \\
\multicolumn{4}{l}{} & \multicolumn{4}{c}{PSNR} & \multicolumn{4}{c}{SSIM} \\ \hline
\multicolumn{4}{l}{VariantA} & \multicolumn{4}{c}{25.97} & \multicolumn{4}{c}{0.757} \\
\multicolumn{4}{l}{VariantB} & \multicolumn{4}{c}{23.47} & \multicolumn{4}{c}{0.698} \\
\multicolumn{4}{l}{VariantC} & \multicolumn{4}{c}{25.44} & \multicolumn{4}{c}{0.729} \\
\multicolumn{4}{l}{VariantD} & \multicolumn{4}{c}{25.81} & \multicolumn{4}{c}{0.746} \\
\multicolumn{4}{l}{\textbf{VariantE (Proposed)}} & \multicolumn{4}{c}{\textbf{26.83}} & \multicolumn{4}{c}{\textbf{0.789}} \\ \hline \hline

\end{tabular}
\end{table*}

\subsection{Effect of UBCDTN}
This experiment is conducted by VariantA and VariantE. Specifically, because of removing UBCDTN, VariantA is trained on bicubic downsampled LR images directly while VariantE takes real-like LR images obtained by UBCDTN as the training LR inputs. According to the analysis of the performance between VariantA and VariantE, we can demonstrate advantages originating from UBCDTN. As shown in Fig. 5, VariantA produces over-smoothed SR images missing high-frequency details while the SR results of VariantE contain naturally desired edges and textures. In addition, from Table III, the quantitative results of VariantA decrease dramatically from 26.83dB/0.789 to 25.97dB/0.757 after removing UBCDTN. The reason is that VariantA simply trains on bicubic data, ignoring the domain distribution difference between bicubic data and real-world data when solving the real-world SR task. By incorporating the UBCDTN in the variant, there is a noteworthy improvement in terms of both qualitative and quantitative performance, which is able to verify that UBCDTN plays an important role in the super-resolution procedure. Thus, we can validate the effectiveness of the proposed UBCDTN and its necessity.

\subsection{Effect of $G_{B}$ and $D_{A}$}
In this experiment, we compare VariantB and VariantE to verify the effect of $G_{B}$ and $D_{A}$, which is also identical to show the effectiveness of the backward cycle module. Note that VariantB is equivalent to UBCDTN for domain transformation without the backward cycle module. In this setting, VariantB takes the artificially degraded images as the input to $G_{A}$ and generates real-like LR images through the supervision of $D_{B}$ in the forward cycle module without the backward cycle module. It can be observed from Table III that with the absence of the backward cycle module, VariantB performs worse than VariantE which contains the whole UBCDTN in terms of PSNR/SSIM values, since there is no restriction to prevent the forward cycle module and backward cycle module from contradicting each other. A huge enhancement can be observed after integrating the backward cycle module into VariantE, where it is able to greatly improve quantitative performance and further produce high quality SR images with desirable details. This is due to the presence of a backward cycle module. The variant is capable of utilizing cycle consistency constraint, which guarantees the correction between two inverse modules in an unsupervised manner. In a word, by introducing $G_{B}$ and $D_{A}$, the whole bi-directional cycle consistency learning strategy can be established to produce real-like LR images which maintain the same characteristics as real-world LR images. These results validate the significance of involving $G_{B}$ and $D_{A}$, which also reveals that the backward cycle module can improve quantitative results and visual quality.

\subsection{Effect of $D_{A}$ and $D_{B}$}
In this experiment, we aim to demonstrate the contribution of $D_{A}$ and $D_{B}$, which can also reflect the importance of an adversarial loss in the UBCDTN. The VariantC and VariantE are compared  to investigate the effect of $D_{A}$ and $D_{B}$. In this case, VariantC is trained on cycle-consistency loss, identity loss, and cycle-perceptual loss, while the adversarial loss is not involved resulting from removing $D_{A}$ and $D_{B}$. As shown in Table III, it is obvious that performance severely decreases by removing discriminators from VariantC. By incorporating $D_{A}$ and $D_B$ into VariantE, we observe that VariantE significantly outperforms VariantC by a large margin of 1.39dB/0.06 in terms of PSNR/SSIM. From Fig. 5, the visual results of VariantC are degraded, where the SR images contain over-smoothed textures and unclear artifacts, while VarantE is able to generate visually realistic SR images with more nature-looking details. The underlying reason for poor performance is that we cannot employ adversarial loss on VariantC, which leads to the lack of adequate iterative adversarial training. According to the quantitative and qualitative comparisons, it can be verified that by applying adversarial loss performed by $D_{A}$ and $D_{B}$ to the variant, the SR performance is greatly enhanced, indicating the effectiveness of $D_{A}$, $D_{B}$ and adversarial loss. 

\subsection{Effect of $FE_{A}$ and $FE_{B}$}
We compare VariantD and VariantE in this experiment to verify the effectiveness of $FE_{A}$ and $FE_{B}$, where the advantages of $FE_{A}$ and $FE_{B}$ are also equal to the benefit of cycle-perceptual loss. By removing $FE_{A}$ and $FE_{B}$, VariantD no longer employs cycle-perceptual loss on the training phase to optimize the model. As shown in Table III, VariantE achieves higher quantitative performance compared to VariantD, increasing PSNR/SSIM values from 25.81dB/0.746 to 26.83dB/0.789. In addition, Fig. 5 clearly shows the qualitative comparisons evaluated by the SR results from VariantD and VariantE with and without cycle-perceptual loss in the training process. It is apparent that without cycle-perceptual loss, the deteriorated textures are visible in the SR results of VariantD. In contrast, when integrating $FE_{A}$ and $FE_{B}$ into the variant, the cycle-perceptual loss can be calculated, which further motivates the variant to produce perceptually natural images with exceptional details. According to the numerical performance and visual results, we are able to verify that $FE_{A}$ and $FE_{B}$ have a significant impact on the super-resolution process, which also identifies that the cycle-perceptual loss is able to improve the perceptual quality of SR images.

\subsection{Final Effect}
The VariantE can be considered as the ultimate proposed method, which includes all the proposed components. Compared with other variants, the ultimate proposed method is able to greatly improve quantitative performance and obviously enhance the quality of visual results. Thus, we can conclude the effectiveness of the proposed method as well as all the components.

\section{Conclusion}
We proposed an unsupervised super-resolution method UBCDTL-GAN for real-world scenarios, which does not involve any paired image data and pre-defined degradation operation. The proposed method comprises two networks, UBCDTN and SESRN. First, the UBCDTN transfers an artificially degraded image to a real-like image with real-world artifacts and characteristics. Next, SESRN reconstructs from the approximate real-like LR image to a visually pleasant super-resolved image with realistic details and textures. According to the designed framework and applied optimization constraints, the proposed method UBCDTL-GAN has the ability to improve real-world super-resolution performance. The quantitative and qualitative experiments on NTIRE 2020 T1 real-world SR dataset validate the effectiveness of our method and show superior SR performances compared to existing state-of-the-art methods.

{\large
\bibliographystyle{IEEEtran}
\bibliography{IEEEabrv}}

\end{document}